\def\year{2020}\relax
\documentclass[letterpaper]{article} % DO NOT CHANGE THIS
\usepackage{aaai20}  % DO NOT CHANGE THIS
\usepackage{times}  % DO NOT CHANGE THIS
\usepackage{helvet} % DO NOT CHANGE THIS
\usepackage{courier}  % DO NOT CHANGE THIS
\usepackage[hyphens]{url}  % DO NOT CHANGE THIS
\usepackage{graphicx} % DO NOT CHANGE THIS
\usepackage{graphicx}  % DO NOT CHANGE THIS
\usepackage{amsmath}
\usepackage{amsfonts}
\usepackage{multirow}
\usepackage{setspace}
\usepackage{bibspacing}
\usepackage[colorlinks,
            linkcolor=black,
            anchorcolor=black,
            citecolor=black]{hyperref}

\def\lbf{{\boldsymbol\ell}}
\def\lnf{{\ell}}
\def\gou{{$\checkmark$}}

\title{Spatial-Temporal Multi-Cue Network for Continuous \\ Sign Language Recognition}
\author{Hao Zhou,\quad Wengang Zhou,\quad Yun Zhou,\quad Houqiang Li\\
CAS Key Laboratory of GIPAS, University of Science and Technology of China \\
zhouh156@mail.ustc.edu.cn,\quad \{zhwg, zhouyun, lihq\}@ustc.edu.cn\\
}
 
\begin{document}

\maketitle

\begin{abstract}
Despite the recent success of deep learning in continuous sign language recognition (CSLR), deep models typically focus on the most discriminative features, ignoring other potentially non-trivial and informative contents. 
Such characteristic heavily constrains their capability to learn implicit visual grammars behind the collaboration of different visual cues (\emph{i,e.}, hand shape, facial expression and body posture). 
By injecting multi-cue learning into neural network design, 
we propose a spatial-temporal multi-cue (STMC) network to solve the vision-based sequence learning problem. 
Our STMC network consists of a spatial multi-cue (SMC) module and a temporal multi-cue (TMC) module. 
The SMC module is dedicated to spatial representation and explicitly decomposes visual features of different cues with the aid of a self-contained pose estimation branch. 
The TMC module models temporal correlations along two parallel paths, \emph{i.e.}, intra-cue and inter-cue, 
which aims to preserve the uniqueness and explore the collaboration of multiple cues. 
Finally, we design a joint optimization strategy to achieve the end-to-end sequence learning of the STMC network. 
To validate the effectiveness,  
we perform experiments on three large-scale CSLR benchmarks:
PHOENIX-2014, CSL and PHOENIX-2014-T. 
Experimental results demonstrate that the proposed method achieves new state-of-the-art performance on all three benchmarks.
\end{abstract}

\section{Introduction}
Sign language is the primary language of the deaf community. 
To facilitate the daily communication between the deaf-mute and the hearing people, it is significant to develop sign language recognition (SLR) techniques. 
Recently, SLR has gained considerable attention for its abundant visual information and systematic grammar rules~\cite{staged,han,tpami19,ian,aaai2019skeleton}. 
In this paper, we concentrate on continuous SLR (CSLR), which aims to translate a series of signs to the corresponding sign gloss sentence.

Sign language mainly relies on, but not limits to, hand gestures. 
To effectively and accurately express the desired idea, sign language simultaneously leverages both manual elements from hands and non-manual elements from the face and upper-body posture~\cite{phoenixdataset2014}. 
To be specific, manual elements include the shape, position, orientation and movement of both hands, 
while non-manual elements include the eye gaze, mouth shape, facial expression and body pose. 
Human visual perception allows us to process and analyze these simultaneous yet complex information without much effort. 
However, with no expert knowledge, it is difficult for a deep neural network to discover the implicit collaboration of multiple visual cues automatically. 
Especially for CSLR, the transitions between sign glosses may come with temporal variations and switches of different cues.

To explore multi-cue information, some methods rely on external tools. 
For example, an off-the-shelf detector is utilized for hand detection, together with a tracker to cope with shape variation and occlusion~\cite{subunet,han}. 
Some methods adopt multi-stream networks with inferred labels (\emph{i.e.}, mouth shape labels, hand shape labels) to guide each stream to focus on individual visual cue~\cite{tpami19}. 
Despite their improvement, they mostly suffer two limitations: 
First, external tools impede the end-to-end learning on the differentiable structure of neural networks. 
Second, off-the-shelf tools and multi-stream networks bring repetitive feature extraction of the same region, incurring expensive computational overhead for such a video-based translation task.

To temporally exploit multi-cue features, an intuitive idea is to concatenate features and feed them into a temporal fusion module. In action recognition, two-stream fusion shows significant performance improvement by fusing temporal features of RGB and optical flow~\cite{vgg-two-stream,vgg-two-stream-2}. 
Nevertheless, the aforementioned fusion approaches are based on two counterpart features in terms of the representation capability. 
But when it turns to multiple diverse cues with unequal feature importance, how to fully exploit the synergy between strong features and weak features still leaves a challenge. 
Moreover, for deep learning based methods, neural networks tend to merely focus on strong features for quick convergence, potentially omitting other informative cues, which limits the further performance improvement.

To address the above difficulties, we propose a novel spatial-temporal multi-cue (STMC) framework. 
In the SMC module, we add two extra deconvolutional layers~\cite{deconv,xiao2018simple} for pose estimation on the top layer of our backbone. 
A soft-argmax trick~\cite{soft-argmax} is utilized to regress the positions of keypoints and make it differentiable for subsequent operations in the temporal part. 
The spatial representations of other cues are acquired by the reuse of feature maps from the middle layer. 
Based on the learned spatial representations, we decompose the temporal modelling part into the intra-cue path and inter-cue path in the TMC module. 
The inter-cue path fuses the temporal correlations between different cues with temporal convolutional (TCOV) layers. 
The intra-cue path models the internal temporal dependency of each cue and feeds them to the inter-cue path at different time scales.  
To fully exploit the potential of STMC network, we design a joint optimization strategy with connectionist temporal classification (CTC)~\cite{CTC} and keypoint regression, making the whole structure end-to-end trainable.

Our main contributions are summarized as follows:
\begin{itemize}
\item We design an SMC module with a self-contained pose estimation branch. It provides multi-cue features in an end-to-end fashion and maintains efficiency at the same time.
\item We propose a TMC module composed of stacked TMC blocks. Each block includes intra-cue and inter-cue paths to preserve the uniqueness and explore the synergy of different cues at the same time.
\item A joint optimization strategy is proposed for the end-to-end sequence learning of our STMC network. 
\item Through extensive experiments, we demonstrate that our STMC network surpasses previous state-of-the-art models on three publicly available CSLR benchmarks. 
\end{itemize}

\section{Related Work}
In this section, we briefly review the related work on sign language recognition and multi-cue fusion.

A CSLR system usually consists of two parts: video representation and sequence learning. Early works utilize 
hand-crafted features~\cite{cooper2009learning,buehler2009learning,yin2016iterative} for SLR. 
Recently, deep learning based methods have been applied to SLR for their strong representation capability. 2D 
convolutional neural networks (2D-CNN) and 3D convolutional neural networks (3D-CNN)~\cite{yukai3d,qiu2017learning-p3d} are employed for modelling the appearance and motion in sign language videos. In~\cite{staged}, Cui \textit{et al.} propose to combine 2D-CNN with temporal convolutional layers for spatial-temporal representation. In~\cite{molchanov2016online,dilated,zhou2019dynamic,wei2019bigmm}, 3D-CNN is adopted to learn motion features in sign language.

Sequence learning in CSLR is to learn the correspondence between video sequence and sign gloss sequence. 
Koller \textit{et al.}~\cite{deephand,resign,koller2018ijcv} propose to integrate 2D-CNNs with hidden markov models (HMM) to model the state transitions. 
In~\cite{subunet,MM,staged,cui-tmm19}, connectionist temporal classification (CTC)~\cite{CTC} algorithm is employed as a cost function for CSLR, which is able to process unsegmented input data. 
In~\cite{han,hlstm}, the attention-based encoder-decoder model~\cite{seq2seq} is adopted to deal with CSLR in the way of neural machine translation. 

The multiple cues of sign language can be separated into categories of multi-modality and multi-semantic. 
Early works about multi-modality utilize physical sensors to collect the 3D space information, such as depth and infrared maps~\cite{molchanov2016online,liu2017continuous}. 
With the development of flow estimation, Cui \textit{et al.}~\cite{cui-tmm19} explore the multi-modality fusion of RGB and optical flow and achieve state-of-the-art performance on PHOENIX-2014 database. 
In contrast, multi-semantic refers to human body parts with different semantics.
Early works use hand-crafted features from segmented hands, tracked body-parts and trajectories for recognition~\cite{buehler2009learning,pfister2013large,phoenixdataset2014}. 
In~\cite{subunet,han}, feature sequence of hand patches captured by a tracker is fused with feature sequence of full-frames for further sequence prediction. 
In~\cite{tpami19}, Koller \textit{et al.} propose to infer weak mouth labels from spoken German annotations and weak hand labels from SL dictionaries. These weak labels are used to establish the state synchronization in HMM of different cues, including full-frame, hand shape and mouth shape. 
Unlike previous methods, we propose an end-to-end differentiable network for multi-cue fusion with joint optimization, which achieves excellent performance. 

\begin{figure*}[ht]
    \centering
    \includegraphics[width=1.0\textwidth]{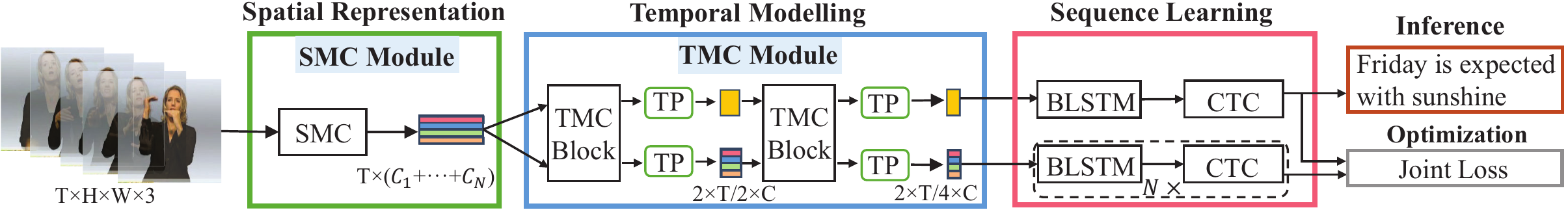}
    \caption{An overview of the proposed STMC framework. The SMC module is firstly utilized to decompose spatial features of visual cues for $T$ frames in a video. Strips with different colors represent feature sequences of different cues. Then, the feature sequences of cues are fed into the TMC module with stacked TMC blocks and temporal pooling (TP) layers. The output of TMC module consists of feature sequence in the inter-cue path and feature sequences of $N$ cues in the intra-cue path, which are processed by BLSTM encoders and CTC layers for sequence learning and inference. Here, $N$ denotes the number of cues.}\label{fig:stmc_overview}
\end{figure*}

\section{Proposed Approach}
In this section, we first introduce the overall architecture of the proposed method. Then we elaborate the key components in our framework, including the spatial multi-cue (SMC) module and temporal multi-cue (TMC) module. Finally, we detail the sequence learning part and the joint loss optimization of our spatial-temporal multi-cue (STMC) framework. 

\subsection{Framework Overview}
Given a video $\mathbf{x}=\{x_t\}_{t=1}^T$ with $T$ frames, the target of CSLR task is to predict its corresponding sign gloss sequence $\lbf=\{\lnf_i\}_{i=1}^L$ with $L$ words. 
As illustrated in Figure~\ref{fig:stmc_overview}, our framework consists of three key modules, \emph{i.e.}, spatial representation, temporal modelling and sequence learning.  
First, each frame is processed by an SMC module to generate spatial features of multiple cues, including full-frame, hand, face and pose. Then, a TMC module is leveraged to capture the temporal correlations of intra-cue features and inter-cue features at different time steps and time scales. Finally, the whole STMC network equipped with bidirectional Long-Short Term Memory (BLSTM)~\cite{hochreiter1997long} encoders utilizes connectionist temporal classification (CTC) for sequence learning and inference.

\begin{figure}[t] 
    \centering
    \includegraphics[width=0.470\textwidth]{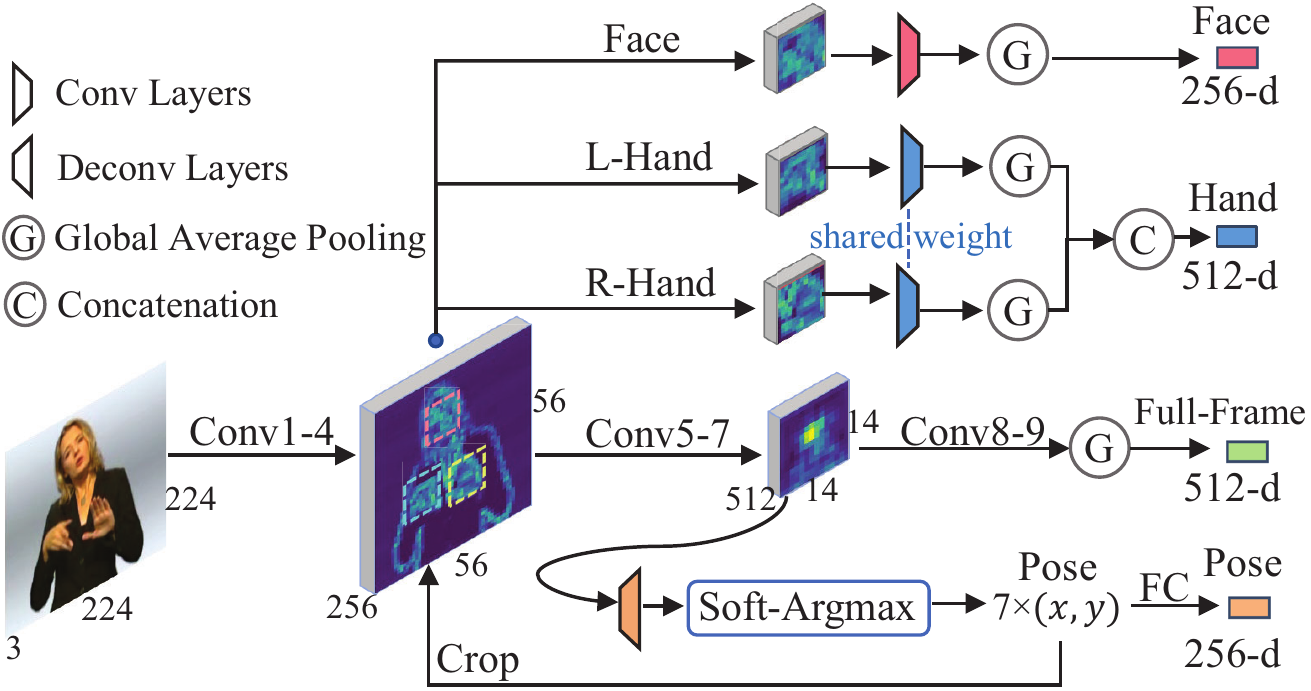}
    \caption{The SMC Module. The keypoints are estimated for patch cropping of face and hands. The output of SMC includes features from full-frame, hands, face and pose.} \label{fig:smc}
\end{figure}

\subsection{Spatial Multi-Cue Representation}
In spatial representation module, 2D-CNN is adopted to generate multi-cue features of full-frame, hands, face and pose. Here, we select VGG-11 model~\cite{vgg} as the backbone network, considering its simple but effective neural architecture design. As depicted in Figure~\ref{fig:smc}, the operations in SMC are composed of three steps: pose estimation, patch cropping and feature generation. 

\subsubsection{Pose Estimation.}
Deconvolutional networks~\cite{deconv} are widely used in pixel-wise prediction. For pose estimation, two deconvolutional layers are added after the 7-th convolutional layer of VGG-11. The stride of each layer is 2. So, the feature maps are $4\times$ upsampled from the resolution $14\times14$ to $56\times56$. The output is fed into a point-wise convolutional layer to generate $K$ predicted heat maps. In each heat map, the position of its corresponding keypoint is expected to show the highest response value. Here, $K$ is set to 7 for keypoints at the upper body, including the nose, both shoulders, both elbows and both wrists. 

To make the keypoint prediction differentiable for subsequent sequence learning, a soft-argmax layer is applied on these heat maps. Denoting $K$ heat maps as $\mathbf{h}=\{h_k\}_{k=1}^K$ , each heat map $h_k\in \mathbb{R}^{H\times W}$ is passed through a spatial softmax function as follows, 
\begin{equation} 
    p_{i,j,k}=\frac{e^{h_{i,j,k}}}{\sum_{i=1}^{H}{\sum_{j=1}^{W}{ e^{h_{i,j,k}} }} },
\end{equation}
where $h_{i,j,k}$ is the value of heat map $h_k$ at position $(i,j)$ and $p_{i, j, k}$ is the probability of keypoint $k$ at position $(i, j)$. Afterwards, the expected values of coordinates along x-axis and y-axis over the whole probability map are calculated as follows,
\begin{equation}
    (\hat{x}, \hat{y})_k \!=\! \left(\sum_{i=1}^{H}{\sum_{j=1}^{W}{ \frac{i\!-\!1}{H\!-\!1} p_{i,j,k} }}, \sum_{i=1}^{H}{\sum_{j=1}^{W}{ \frac{j\!-\!1}{W\!-\!1} p_{i,j,k} }}\right).
\end{equation}
Here, $J_k\!=\!(\hat{x},\hat{y})_k\in[0,1]$ is the normalized predicted position of keypoint $k$. The corresponding position of $(x, y)$ in a $H\times W$ feature map is $\left(\hat{x}(H\!-\!1)+1, \hat{y}(W\!-\!1)+1 \right)$.

\subsubsection{Patch Cropping.}
In CSLR, the perception of detailed visual cues is vital, including eye gaze, facial expression, mouth shape, hand shape and orientations of hands. 
Our model takes predicted positions of the nose and both wrists as the center points of the face and both hands. 
The patches are cropped from the output ($56\times 56\times C_4$) of 4-th convolutional layer of VGG-11. The cropping sizes are fixed to $24\times 24$ for both hands and $16\times16$ for the face. It's large enough to cover body parts of a signer whose upper body is visible to the camera. The center point of each patch is clamped into a range to ensure that the patch would not cross the border of the original feature map.

\subsubsection{Feature Generation.}
After $K$ keypoints are predicted, they are flattened to a 1D-vector with dimension $2K$ and passed through two fully-connected (FC) layers with ReLU to get the feature vector of pose cue. Then, feature maps of the face and both hands are cropped and processed by several convolutional layers, separately. Most sign gestures rely on the cooperation of both hands. So we use weight-sharing convolutional layers for both hands. The outputs of them are concatenated along the channel-dimension. Finally, we perform global average pooling over all the feature maps with spatial dimension to form feature vectors of different cues.

All features are extracted by passing frames $\mathbf{x}=\{x_t\}_{t=1}^T$ through our spatial multi-cue (SMC) module as follows,
\begin{equation} \label{eq:smc}
    \left\{ \left\{f_{t,n}\right\}_{n=1}^N , \left\{J_{t,k}\right\}_{k=1}^K \right\}_{t=1}^T = \left\{ \Omega_\theta(x_t) \right\}_{t=1}^T,
\end{equation}
where \(\Omega_\theta(\cdot)\) denotes SMC module and $\theta$ denotes the parameters of it. $J_{t,k}\in\mathbb{R}^2$ is the position of keypoint $k$ at the $t$-th frame. $f_{t,n}\in\mathbb{R}^{C_n}$ is the feature vector of visual cue $n$ at the $t$-th frame. In this paper, we set $N=4$, which represents visual cues of full-frame, hand, face and pose, respectively.

\subsection{Temporal Multi-Cue Modelling}
Instead of simple fusion, our proposed temporal multi-cue (TMC) module intends to integrate spatiotemporal information from two aspects, intra-cue and inter-cue. The intra-cue path captures the unique features of each visual cue. The inter-cue path learns the combination of fused features from different cues at different time scales. Then, we define a TMC block to model the operations between the two paths as follows,
\begin{equation}
    ({o_l,f_l}) = {\rm Block}_l({o_{l-1}, f_{l-1}}),
\end{equation}
where $(o_{l-1},f_{l-1})$ and $(o_l,f_l)$ are the input pair and output pair of the $l$-th block. 
$o_l\in\mathbb{R}^{T\times C_o}$ denotes the feature matrix of the inter-cue path. 
$f_l\in\mathbb{R}^{T\times C_f}$ denotes the feature matrix of the intra-cue path which is the concatenation of vectors from different cues along channel-dimension. 
As the first input pair, $o_1\!=\!f_1\!=\![f_{1,1}, f_{1,2}, \cdots, f_{1,N}]$, where $[\cdot]$ is the concatenating operation and $N$ is the number of cues. 

\begin{figure}[t]
    \centering
    \includegraphics[width=0.47\textwidth]{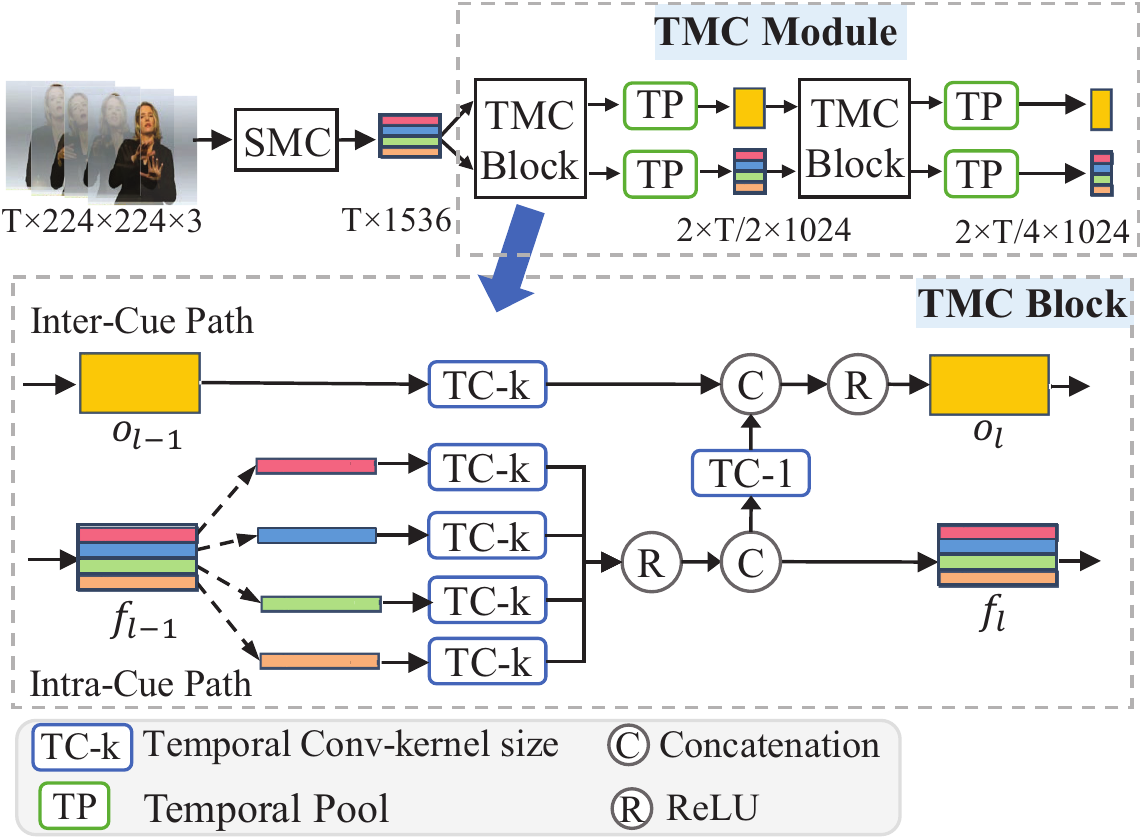}
    \caption{The TMC Module.}\label{fig:tmc}
  \end{figure}

The detailed operations inside each TMC block are shown in Figure~\ref{fig:tmc} and can be decomposed into two paths as follows. ( $C$ is the number of output channels in each path)

\subsubsection{Intra-Cue Path.}
The first path is to provide unique features of different cues at different time scales. 
The temporal transformation inside each cue is performed as follows,
\begin{align}
    f_{l,n} &= {\rm ReLU}( \mathcal{K}_{k}^{\frac{C}{N}}(f_{l-1,n}) ), \\
    f_l &= [f_{l,1}, f_{l,2}, \cdots, f_{l,N}]. 
\end{align}
Here, $f_{l,n}\in\mathbb{R}^{T\times \frac{C}{N}}$ denotes the feature matrix of $n$-th cue. $\mathcal{K}_k^{\frac{C}{N}}$ denotes the kernel of a temporal convolution, where $k$ is the temporal kernel size and $\frac{C}{N}$ is the number of output channels.

\subsubsection{Inter-Cue Path.}
The second path is to perform the temporal transformation on the inter-cue feature from the previous block and fuse information from the intra-cue path as follows,
\begin{equation}
    o_{l} = {\rm ReLU}([\mathcal{K}_k^{\frac{C}{2}}(o_{l-1}), \mathcal{K}_1^{\frac{C}{2}}(f_{l}) ]),
\end{equation}
where $\mathcal{K}_1^{\frac{C}{2}}$ is a point-wise temporal convolution. It serves as a project matrix between the two paths. Note that $f_l$ is the output of intra-cue path in the present block.

After each block, a temporal max-pooling with stride 2 and kernel size 2 is performed. In this paper, we use two blocks in the TMC module. The kernel size $k$ of all temporal convolutions is set to 5, except the point-wise one. The number of output channels $C$ in each path is set to 1024.

\subsection{Sequence Learning and Inference}
With the proposed SMC and TMC module, the network can generate inter-cue feature sequence $\mathbf{o}=\{o_t\}_{t=1}^{T'}$ and $N$ intra-cue feature sequences $\mathbf{f}_n=\{f_{t,n}\}_{t=1}^{T'}$. Here, $T'$ is the temporal length of the final output of the TMC module. The question then is how to utilize these two feature sequences to accomplish the sequence learning and inference.
\subsubsection{BLSTM Encoder.}
Recurrent neural networks (RNN) can use their internal state to model the state transitions in the sequence of inputs. Here, we use RNN to map the spatial-temporal feature sequence to its sign gloss sequence. RNN takes the feature sequence as input and generates $T'$ hidden states as follows,
\begin{equation}
    h_t={\rm RNN}(h_{t-1}, o_t),
\end{equation}  
in which $h_t$ is the hidden state at time step $t$ and the initial state $h_0$ is a fixed all-zero vector. 
In our approach, we choose the bidirectional Long Short-Term Memory (BLSTM)~\cite{LSTM} unit as the recurrent unit for its ability in processing long-term dependencies. BLSTM concatenates forward and backward hidden states from bidirectional inputs. Afterward, the hidden state of each time step is passed through a fully-connected layer and a softmax layer,
\begin{equation}\label{rnn}
    \begin{split}
    a_t = W \cdot h_t+b, \quad
    y_{t,j}=\frac{e^{a_{t,j}}}{\sum_k e^{a_{t,k}}},
    \end{split}
\end{equation}
where $y_{t,j}$ is the probability of label $j$ at time step $t$. In CSLR task, label $j$ comes from a given vocabulary.

\subsubsection{Connectionist Temporal Classification.}
Our model employs connectionist temporal classification (CTC)~\cite{CTC} to tackle the problem of mapping video sequence $\mathbf{o}=\{o_t\}_{t=1}^{T'}$ to ordered sign gloss sequence $\lbf=\{\lnf_i\}_{i=1}^L$ ($L \le T$), where the explicit alignment between them is unknown. 
The objective of CTC is to maximize the sum of probabilities of all possible alignment paths between input and target sequence.

CTC creates an extended vocabulary $\mathcal{V}$ with a blank label ``\(-\)'', where $\mathcal{V}=\mathcal{V}_\text{origin} \cup \{-\}$. The blank label represents stillness and transitions which have no precise meaning. 
Denote the alignment path of the input sequence as $\pi = \{ \pi_t \}_{t=1}^{T'}$, where label $\pi_t \in \mathcal{V}$. The probability of alignment path $\pi$ given the input sequence is defined as follows,
\begin{equation}
    p(\pi | \mathbf{o}) = \prod_{t=1}^{T'} p(\pi_t | \mathbf{o})= \prod_{t=1}^{T'} y_{t,\pi_t}.
\end{equation}

Define a many-to-one mapping operation $\mathcal{B}$ which removes all blanks and repeated words in the alignment path (\emph{e.g.}, $\mathcal{B}(II-miss--you)=I,miss,you$). In this way, we calculate the conditional probability of sign gloss sequence $\lbf$ as the sum of probabilities of all paths that can be mapped to $\lbf$ via $\mathcal{B}$:
\begin{equation}
    p(\lbf | \mathbf{o}) = \sum_{\pi \in \mathcal{B}^{-1}(\lbf)}  p(\pi | \mathbf{o}),
\end{equation}
where \(\mathcal{B}^{-1}(\lbf)\!=\!\{\pi | \mathcal{B}(\pi)\!=\!\lbf\}\) is the inverse operation of \(\mathcal{B}\). Finally, the CTC losses of inter-cue feature sequence $\mathbf{o}$ and intra-cue feature sequence $\mathbf{f_n}$ are defined as follows,
\begin{align}
    &\mathcal{L}_{\text{CTC}-\mathbf{o}} = -\ln{p(\lbf | \mathbf{o})}, \\
    &\mathcal{L}_{\text{CTC}-\mathbf{f}_n} = -\ln{p(\lbf | \mathbf{f}_n)}.
\end{align}

\subsubsection{Joint Loss Optimization.}
During the training process, we take the optimization of the inter-cue path as the primary objective. To provide the information of each individual cue for fusion, the intra-cue path plays an auxiliary role. Hence, the objective function of the entire STMC framework is given as follows,
\begin{equation} \label{eq:joint_loss}
    \mathcal{L} = \mathcal{L}_{\text{CTC}-\mathbf{o}}+\alpha \sum\nolimits_{n}{\mathcal{L}_{\text{CTC}-\mathbf{f}_n}} + \mathcal{L}_{\text{R}}^{\beta}.
\end{equation} 
Here, $\alpha$ and $\beta$ are hyper-parameters, where $\alpha$ is to balance the ratio of auxiliary loss for the intra-cue path, and $\beta$ is to make the regression loss $\mathcal{L}_{\text{R}}$ of pose estimation have the same order of magnitudes with others. 
Given the estimated keypoints $J_{t,k}\in\mathbb{R}^2$ which is calculated in Eq.~\ref{eq:smc}, its corresponding ground-truth is $\hat{J}$, and the smooth-L1~\cite{girshick2015fast} loss function of pose estimation branch is calculated as follows,
\begin{equation}
    \mathcal{L}_{\text{R}}^{\beta} = \frac{1}{2T\!K} \sum_t \sum_k \sum_{i\in(x,y)}  \text{smooth}_{L_1}\beta( J_{t,k,i}- \hat{J}_{t,k,i}),
\end{equation}
in which,
\begin{align}
    \text{smooth}_{L_1}(x) = \begin{cases}
          0.5 x^2  &\text{if $|x|<1$,} \\
          |x|-0.5  &\text{otherwise.}      
        \end{cases} 
\end{align}

\subsubsection{Inference.}
For inference, we pass video frames through the SMC and TMC modules. Only the inter-cue feature sequence and its BLSTM encoder are used to generate the posterior probability distribution of glosses at all time steps. We use the beam search decoder~\cite{deepspeech} to search the most probable sequence within an acceptable range.

\section{Experiments}
\subsection{Dataset and Evaluation}
\subsubsection{Dataset.}
We evaluate our method on three datasets, including PHOENIX-2014~\cite{phoenixdataset2014}, CSL~\cite{han,hlstm} and PHOENIX-2014-T~\cite{phoenix2014t}. 

\textbf{PHOENIX-2014} is a publicly available German Sign Language dataset, which is the most popular benchmark for CSLR. The corpus was recorded from broadcast news about the weather. It contains videos of 9 different signers with a vocabulary size of 1295. The split of videos for Train, Dev and Test is 5672, 540 and 629, respectively. Our method is evaluated on the multi-signer database.

\textbf{CSL} is a Chinese Sign Language dataset, which has 100 sign language sentences about daily life with 178 words. Each sentence is performed by 50 signers and there are 5000 videos in total. For pre-training, it also provides a matched isolated Chinese Sign Language database,  
which contains 500 words. Each word is performed 10 times by 50 signers. 

\textbf{PHOENIX-2014-T} is an extended version of PHOENIX-2014 and has two-stage annotations for new videos.
One is sign gloss annotations for CSLR task. Another is German translation annotations for sign language translation (SLT) task. The split of videos for Train, Dev and Test is 7096, 519 and 642, respectively. It has no overlap with the previous version between Train, Dev and Test set. The vocabulary size is 1115 for sign gloss and 3000 for German. 

\subsubsection{Pose Annotation.}
To obtain the keypoint positions for training, we use the publicly available HRNet~\cite{sun2019deep} toolbox to estimate the positions of 7 keypoints in upper-body for all frames on three databases. The toolbox gives 2D coordinates $(x,y)$ in the pixel coordinate system. We thus represent each normalized keypoint with a tuple of $(x,y)$ and record it as an array of 7 tuples.

\subsubsection{Evaluation.}
In CSLR, Word Error Rate (WER) is used as the metric of measuring the similarity between two sentences~\cite{phoenixdataset2014}. It measures the least operations of substitution (sub), deletion (del) and insertion (ins) to transform the hypothesis to the reference:
\begin{equation}
\text{WER} = \frac{\text{\#substitutions \,+\, \#deletions \,+\, \#insertions}} {\text{\#words in reference}}.
\end{equation}

\begin{figure}[t]
    \centering
    \includegraphics[width=0.47\textwidth]{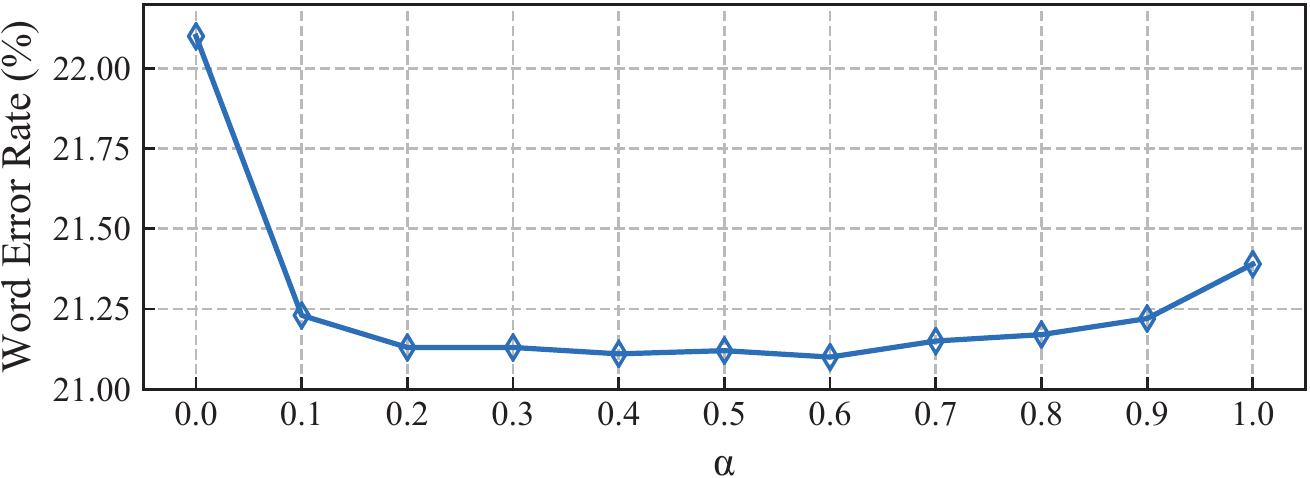}
    \caption{The effect of weight parameter $\alpha$ in Eq.~\ref{eq:joint_loss}.}\label{fig:alpha}
\end{figure}

\begin{figure*}[ht] 
    \centering
    \includegraphics[width=\textwidth]{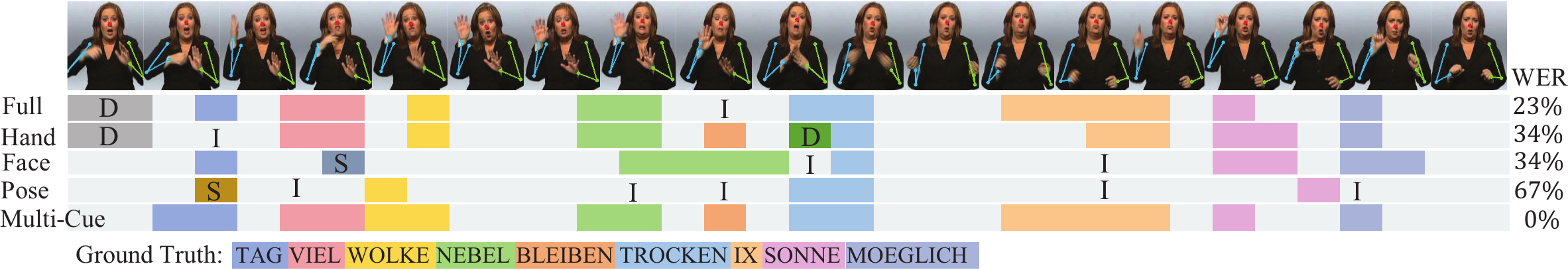}
    \caption{A qualitative result of different cues with estimated poses (zoom in) from Dev set (D: delete, I: insert, S: substitute).} \label{fig:demo}
\end{figure*}

\subsection{Implementation Details}
In our experiments, the input frames are resized to $224\times224$. 
For data augmentation in one video, we add random crop at the same location of all frames, random discard of $20\%$ frames and random flip of all frames. 
For inter-cue features, the number of output channels after TCOVs and BLSTM are all set to 1024. There are 4 visual cues. For each intra-cue feature, the number of output channels after TCOVs and BLSTM are all set to 256. 

Following the previous methods~\cite{resign,ian,cui-tmm19}, we utilize a staged optimization strategy. First, we train a VGG11-based network as DNF~\cite{cui-tmm19} and use it to decode pseudo labels for each clip. Then, we add a fully-connected layer after each output of the TMC module. The STMC network without BLSTM is trained with cross-entropy and smooth-L1 loss by SGD optimizer. The batch size is 24 and the clip size is 16. Finally, with fine-tuned parameters from the previous stage, our full STMC network is trained end-to-end under joint loss optimization. We use Adam optimizer with learning rate $5\times10^{-5}$ and set the batch size to 2. In all experiments, we set $\alpha$ to 0.6 and $\beta$ to 30. In fact, the experiment results are insensitive to the slight change of $\alpha$ (see Fig.~\ref{fig:alpha}), except $\alpha=0$.

Our network architecture is implemented in PyTorch. For finetuning, we train the STMC network without BLSTM for 25 epochs. Afterward, the whole STMC network is trained end-to-end for 30 epochs. For inference, the beam width is set to 20. Experiments are run on 4 GTX 1080Ti GPUs.

\subsection{Framework Effectiveness Study}
\begin{table}[tp]
    \centering
    \small
    \caption{Evaluation of different module combinations (the lower the better).} \label{tab:module}
    \begin{tabular}{l|cc|cc}
    \hline
    \multirow{2}{*}{Method} & \multicolumn{2}{c|}{Dev} & \multicolumn{2}{c}{Test} \\
                            & del/ins      & WER       & del/ins       & WER      \\ \hline
    VGG+1D-CNN              & 9.0/3.0      & 25.1      & 8.4/3.0       & 25.5     \\
    SMC+1D-CNN              & 7.6/3.8      & 22.7      & 7.4/3.5       & 22.4     \\
    SMC+TMC w/o JL          & 6.6/4.7      & 22.1      & 6.7/3.8       & 21.9     \\
    SMC+TMC                 & 7.7/3.4      & 21.1      & 7.4/2.6       & 20.7     \\ \hline
    \end{tabular}
\end{table}

\begin{table}[tp]
    \centering
    \small
    \caption{Evaluation of different paths in TMC on Dev set.} \label{tab:tmc}
    \begin{tabular*}{0.43\textwidth}{c|@{\extracolsep{\fill}}c|cccc}
    \hline
    \multirow{2}{*}{Path}&  \multirow{2}{*}{Inter-Cue} & \multicolumn{4}{c}{Intra-Cue}  \\ 
           &               & full     & hand   & face & pose \\ \hline
    WER &     21.1         & 25.0     & 30.5   & 35.0 & 51.4   \\ \hline
    \end{tabular*}
\end{table}

\begin{table}[tp]
    \centering
    \small
    \caption{Comparison of inference time. (PE: an external VGG11-based model for pose estimation)} \label{tab:efficiency}
    \begin{tabular*}{0.41\textwidth}{l|@{\extracolsep{\fill}}c|c|c}
    \hline
    Method            & FLOPs  & Time      & WER (Dev) \\ \hline 
    VGG+1D-CNN        &  7.5G  & 264ms     & 25.1 \\         
    VGG+TMC+PE        &  17.3G & 628ms     & 21.7 \\          
    SMC+TMC           &  10.3G & 352ms     & 21.1 \\ \hline   
    \end{tabular*}
\end{table}

For a fair comparison, experiments in this subsection are conducted on PHOENIX-2014, which is the most popular dataset in CSLR.
\subsubsection{Module Analysis}
We analyze the effectiveness of each module in our proposed approach. In Table~\ref{tab:module}, different combinations of spatial and temporal modules are evaluated. The baseline model is composed of VGG11 and 1D-CNN with a BLSTM encoder. With the aid of multi-cue features, the SMC module provides about $3\%$ improvement compared with baseline on the test set. However, with no extra guidance, the TMC module doesn't show expected gain by replacing the 1D-CNN. With joint loss optimization, the intra-cue path is guided by CTC loss to learn temporal dependency of each cue and provides $1.6\%$ and $1.7\%$ extra gain on the dev set and test set, compared with 1D-CNN. Compared with the baseline model, our STMC network reduces the WER on the test set by $4.8\%$. 

\subsubsection{Intra-Cue and Inter-Cue Paths}
With further optimization, the BLSTM encoder of each cue in the intra-cue path can also serve as an individual sequence predictor. In Table~\ref{tab:tmc}, WERs of different encoders in both paths are evaluated on the dev set. Among the four cues, the performance of pose is worst. With only the position and orientation of joints in upper-body, it's difficult to distinguish the subtle variations in the appearance of sign gestures. The performance of hand is superior to that of face, while full-frame achieves relatively better performance. By leveraging the synergy of different cues, the inter-cue path shows the lowest WER. 

\subsubsection{Inference Time} 
To clarify the effectiveness of the self-contained pose estimation branch, we evaluate the inference time in Table~\ref{tab:efficiency}. The inference time depends on the video length. In average, it takes around 8 seconds (25FPS) for a sign sentence. For a fair comparison, we evaluate the inference time of 200 frames on a single GPU. Compared with introducing an external VGG-11 based model for pose estimation, our self-contained branch saves around $44\%$ inference time. It's notable that our framework with the self-contained branch still shows slightly better performance than an off-the-shelf model. We argue that the differentiable pose estimation branch plays the role of regularization and then alleviate the overfitting of neural networks. 

\subsubsection{Qualitative Analysis}
Figure~\ref{fig:demo} shows an example generated by different cues. It's clear to see that the result of the inter-cue path can effectively learn correlations of multiple cues and make a better prediction.

\subsection{State-of-the-art Comparison}

\begin{table*}[tp]
    \centering
    \small
    \caption{Comparison with methods on PHOENIX-2014 (the lower the better).} \label{tab:phoenix2014}
    \begin{tabular*}{0.90\textwidth}{p{6.5cm}@{\extracolsep{\fill}}|cccc|cc|cc}
    \hline
    \multirow{2}{*}{Method}         &\multicolumn{4}{c|}{Cue}        & \multicolumn{2}{c|}{Dev} & \multicolumn{2}{c}{Test} \\
                                    & full & hand & face   & pose  & del/ins  & WER  & del/ins      & WER                  \\ \hline
    CMLLR~\cite{phoenixdataset2014} &      & \gou & \gou   & \gou  & 21.8/3.9 & 55.0 & 20.3/4.5 & 53.0      \\
    1-Mio-Hands~\cite{deephand}     &      & \gou & \gou   & \gou  & 16.3/4.6 & 47.1 & 15.2/4.6 & 45.1      \\
    CNN-Hybrid~\cite{deepsign}      &      & \gou &        &       & 12.6/5.1 & 38.3 & 11.1/5.7 & 38.8      \\
    SubUNets~\cite{subunet}         & \gou & \gou &        &       & 14.6/4.0 & 40.8 & 14.3/4.0 & 40.7      \\ 
    Staged-Opt~\cite{staged}        &      & \gou &        &       & 13.7/7.3 & 39.4 & 12.2/7.5 & 38.7      \\ 
    Re-sign~\cite{resign}           & \gou &      &        &       & -        & 27.1 & -        & 26.8      \\ 
    LS-HAN~\cite{han}               & \gou & \gou &        &       & -        & -    & -        & 38.3      \\ 
    Dilated~\cite{dilated}          & \gou &      &        &       & 8.3/4.8  & 38.0 & 7.6/4.8  & 37.3      \\ 
    Hybrid CNN-HMM~\cite{koller2018ijcv} &  &\gou  &       &       & -       & 31.6 & -        & 32.5      \\ 
    IAN~\cite{ian}                  & \gou &      &        &       & 12.9/2.6 & 37.1 & 13.0/2.5 & 36.7      \\ 
    DenseTCN~\cite{densetcn}        & \gou &      &        &       & 10.7/5.1 & 35.9 & 10.5/5.5 & 36.5      \\ 
    CNN-LSTM-HMM~\cite{tpami19}     & \gou & \gou &        &       & -        & 26.0 & -        & 26.0      \\ 
    DNF~\cite{cui-tmm19} (RGB)        & \gou &      &      &       & 7.8/3.5  & 23.8 & 7.8/3.4  & 24.4      \\ 
    DNF~\cite{cui-tmm19} (RGB+Flow)   & \gou &      &      &       & 7.3/3.3  & 23.1 & 6.7/3.3  & 22.9      \\ \hline 
    SMC  (ours)                     & \gou & \gou & \gou   &\gou   & 7.6/3.8  & 22.7 & 7.4/3.5  & 22.4      \\ 
    STMC (ours)                     & \gou & \gou & \gou   &\gou   & 7.7/3.4  & \textbf{21.1} & 7.4/2.6  & \textbf{20.7}      \\ \hline

    \end{tabular*}
\end{table*}

\begin{table}[tp]
    \centering
    \small
    \caption{Comparison with methods on CSL.} \label{tab:CSL}
    \begin{tabular}{l|c|c}
    \hline
    Method                          & Split I & Split II  \\ \hline
    S2VT~\cite{s2vt-v}              & 25.5    & 67.0      \\
    LS-HAN~\cite{han}               & 17.3    &  -        \\
    HLSTM-attn~\cite{hlstm}         & 10.2    & 64.1      \\
    CTF~\cite{MM}                   & 11.2    &  -        \\ 
    CTM~\cite{ctm}                  &  -      & 61.9      \\
    DenseTCN~\cite{densetcn}        & 14.3    & 44.7      \\
    IAN~\cite{ian}                  &  -      & 32.7      \\ \hline
    STMC (ours)                     & \textbf{2.1}     & \textbf{28.6}      \\ \hline
    \end{tabular}
\end{table}

\begin{table}[tp]
    \centering
    \small
    \caption{Comparison with methods on PHOENIX-2014-T. (f: full-frame, m: mouth, h: hand)} \label{tab:phoenix-2014-t}
    \begin{tabular}{l|c|c}
    \hline
    Method                              & Dev & Test  \\ \hline
    1 stream (f)~\cite{tpami19}         & 24.5    & 26.5      \\
    2 stream (f+m)~\cite{tpami19}       & 24.5    & 25.4      \\
    3 stream (f+m+h)~\cite{tpami19}     & 22.1    & 24.1      \\ \hline
    STMC (ours)                         & \textbf{19.6}     & \textbf{21.0}      \\ \hline
    \end{tabular}
\end{table}

\subsubsection{Evaluation on PHOENIX-2014.} In Table~\ref{tab:phoenix2014}, we compare our approach with methods on PHOENIX-2014. 
CMLLR and 1-Mio-Hands belong to traditional HMM-based model with hand-crafted features. In SubUNets and LS-HAN, full-frame features are fused with features of hand patches, which are captured by an external tracker. In CNN-LSTM-HMM, two-stream networks are trained with weak hand labels and sign gloss labels, respectively. Our STMC outperforms two recent multi-cue methods, \emph{i.e.}, LS-HAN and CNN-LSTM-HMM by $17.6\%$ and $5.3\%$. Moreover, compared with DNF which explores the fusion of RGB and optical flow modality, STMC still surpasses this best competitor by $2.2\%$. Based on the RGB modality, we propose a novel STMC framework and achieves $20.7\%$ WER on the test set, a new state-of-the-art result on PHOENIX-2014.

\subsubsection{Evaluation on CSL.} In Table~\ref{tab:CSL}, we evaluate our approach on CSL under two settings. CSL dataset contains a smaller vocabulary compared with PHOENIX-2014. Following the works of ~\cite{han,hlstm}, the dataset is split by two strategies in Table~\ref{tab:CSL}. \textbf{Split I} is a signer independent test: the train and test sets share the same sentences with no overlap of signers. \textbf{Split II} is an unseen sentence test: the train and test sets share the same signers and vocabulary with no overlap of same sentences. Between two settings, \textbf{Split II} is more challenging for that recognizing unseen combinations of words is difficult in CSLR. In IAN, their alignment algorithm of CTC decoder and LSTM decoder shows notable improvement, compared with previous methods. Benefiting from multi-cue learning, our STMC framework outperforms the best competitor on CSL by $4.1\%$ on WER.

\subsubsection{Evaluation on PHOENIX-2014-T.} In Table~\ref{tab:phoenix-2014-t}, we provide a result of our method on PHOENIX-2014-T. As a newly proposed dataset~\cite{phoenix2014t} for sign language translation, PHOENIX-2014-T provides an extended database with sign gloss annotation and spoken German annotation. CNN-LSTM-HMM utilizes spoken German annotation to infer the weak mouth shape labels for each video. It provides results of multi-cue sequential parallelism, including full-frame, hand and mouth. Our method surpasses all three combinations of CNN-LSTM-HMM.

\section{Conclusion}
In this paper, we present a novel multi-cue framework for CSLR, which aims to learn spatial-temporal correlations of visual cues in an end-to-end fashion. In our framework, a spatial multi-cue module is designed with a self-contained pose estimation branch to decompose spatial multi-cue features. Moreover, we propose a temporal multi-cue module composed of the intra-cue and inter-cue paths, which aims to preserve the uniqueness of each cue and explore the synergy of different cues at the same time. A joint optimization strategy is proposed to accomplish multi-cue sequence learning. Extensive experiments on three large-scale CSLR datasets demonstrate the superiority of our STMC framework.

{\bf Acknowledgments.} This work was supported in part to Dr. Wengang Zhou by NSFC under contract No. 61632019 \& No. 61822208 and Youth Innovation Promotion Association CAS (No. 2018497), and in part to Dr. Houqiang Li by NSFC under contract No. 61836011. 

{
\begin{spacing}{0.94}

% \small
\footnotesize
\bibliography{1648.reference}
\bibliographystyle{aaai}

\end{spacing}
}

\end{document}